\documentclass{llncs}

\usepackage{graphicx}
\usepackage{tabularx}
\usepackage{amsmath}
\usepackage[table]{xcolor}
\usepackage{hyperref}

\newcolumntype{Y}{>{\centering\arraybackslash}X}
\newcolumntype{?}{!{\vrule width 1pt}}

\begin{document}

\title{Weakly Supervised Localisation for Fetal Ultrasound Images}

\titlerunning{Weak Localisation for Fetal US Images}

\author{Nicolas Toussaint\inst{1}
\and Bishesh Khanal\inst{1,2}
\and Matthew Sinclair\inst{2}
\and Alberto Gomez\inst{1}
\and Emily Skelton\inst{1}
\and Jacqueline Matthew\inst{1}
\and Julia A. Schnabel\inst{1}}

\authorrunning{Nicolas Toussaint et al.}

% list of authors for the TOC (use if author list has to be modified)
\tocauthor{Nicolas Toussaint, Bishesh Khanal, Julia Schnabel}

\institute{School of Biomedical Engineering and Imaging Sciences, King's College London, United Kingdom \\
\email{nicolas.a.toussaint@kcl.ac.uk}, \\
\and
Department of Computing, Imperial College London, United Kingdom}

\maketitle

\begin{abstract}
This paper addresses the task of detecting and localising fetal anatomical regions in 2D ultrasound images, where only image-level labels are present at training, i.e. without any localisation or segmentation information. We examine the use of convolutional neural network architectures coupled with soft proposal layers. The resulting network simultaneously performs anatomical region detection (classification) and localisation tasks. We generate a proposal map describing the attention of the network for a particular class. The network is trained on 85,500 2D fetal Ultrasound images and their associated labels. Labels correspond to six anatomical regions: head, spine, thorax, abdomen, limbs, and placenta. Detection achieves an average accuracy of 90\% on individual regions, and show that the proposal maps correlate well with relevant anatomical structures. This work presents itself as a powerful and essential step towards subsequent tasks such as fetal position and pose estimation, organ-specific segmentation, or image-guided navigation. Code and additional material is available at \url{https://ntoussaint.github.io/fetalnav}.

\end{abstract}

\section{Introduction}

Ultrasound (US) is the most popular obstetric imaging modality for antenatal detection of fetal abnormalities. A routine US screening examination consists of manually scanning the fetal anatomy, mainly using 2D imaging, selecting a series of standard planes, and measuring biometric data to assess fetal normality. The plane selection process depends on the local/departmental protocol (e.g. FASP~\cite{uk2015fetal} in the UK). Steering the US transducer to obtain these anatomical planes of interest is a challenging task due to the large variability in image orientation and appearance, within an anatomical region as well as within a standard plane~\cite{sarris2012intra}. Recent years have seen significant efforts to detect such planes in US video sequences~\cite{yaqub2015guided,baumgartner2017sononet}. While these methods are extremely valuable, they disregard more than 95\% of the examination images that do not fall into a standard plane category. The remaining images do however contain valuable information about the global fetal anatomy. With that in mind, categorisation of any generic fetal US image in global anatomical regions is of great clinical interest. Such categorisation could for instance provide anatomical context to a subsequent organ specific task. Furthermore, localising general fetal structures could play an important role in the development of navigation systems, and could also be used to better understand and learn the patterns of steering towards specific planes of interests.

\noindent \textbf{Related Work:} Weakly supervised object localisation is a relatively active field of research in Neural Network literature~\cite{oquab2015object}. Recently, Zhu et al. proposed a their Soft Proposal Networks (SPN)~\cite{zhu2017soft}, consisting of a dedicated extra layer attached to any CNN architecture, specifically designed for this task. It was initially inspired by the class activation map (CAM) approach in~\cite{zhou2016learning}. One of the main advantages of SPNs over conventional Region Proposal Networks~\cite{ren2015faster} for instance is that the proposal itself is an objectness confidence, and does not necessitate back propagation at inference time to retrieve saliency. As a consequence, it can be used directly in an end-to-end learning manner: the proposal couples with convolutional activation and evolves with the deep feature learning.

In the context of US images and fetal screening in particular, recent papers focused on classification or detection~\cite{baumgartner2017sononet} of standard planes in US video sequences. In~\cite{baumgartner2017sononet}, the authors detected a standard plane within a real screening session video. They used saliency maps to infer the regions of interest attached to the detected standard plane using back propagation. They however discard the vast majority of acquired images ($\sim 95\%$) as background. In contrast, our work provides semantic level of labels for any arbitrary image. Thus, during scanning, the proposed method provides useful information and context from all images captured in a fetal US examination.

\noindent \textbf{Contribution: } We propose a method to detect and localise fetal anatomical regions applicable to any arbitrary 2D US fetal image within 22-32 week gestational age. The system is based on Convolutional Neural Networks (CNNs) and soft proposal layers~\cite{zhu2017soft} for weakly supervised localisation. It is to our knowledge the first attempt to transfer this technology to free hand 2D US fetal anatomical region localisation. The network is able to detect six separate anatomical regions of the fetal body with high accuracy ($\sim 90\%$), and localise key anatomical structures within an image in real time ($\sim 20\text{Hz}$).

\begin{figure}[!ht]
\centering
\includegraphics[width=.75\textwidth]{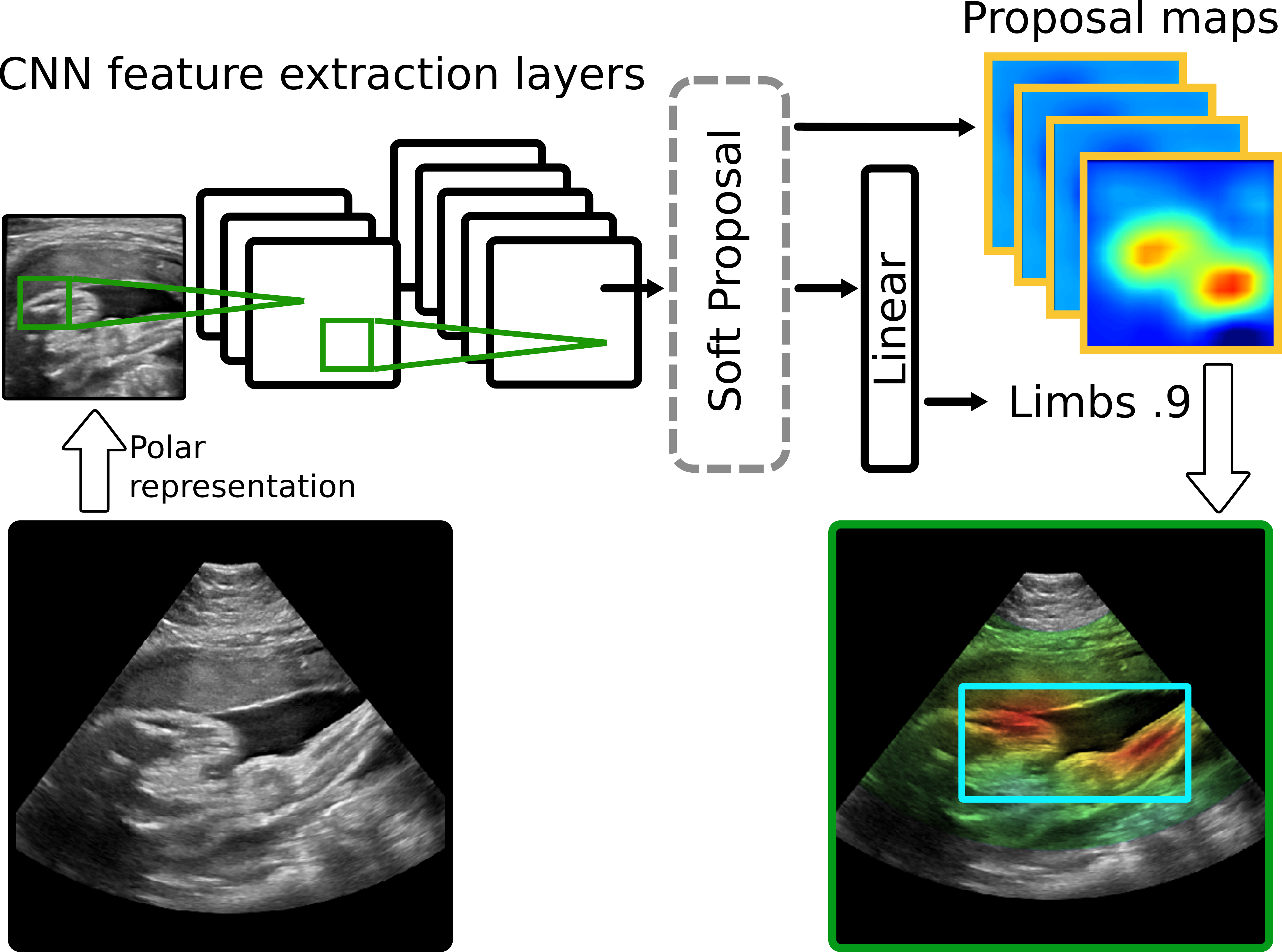}
\caption{Overview of our framework: A fetal ultrasound image is processed in real-time via a Convolutional Neural Network coupled with a Soft Proposal layer. The anatomical region is detected and localised using weakly supervised learning.}
\label{fig:overview}
\end{figure}

\section{Data}
\vspace{-2mm}

\begin{figure}
\includegraphics[width=0.59\textwidth]{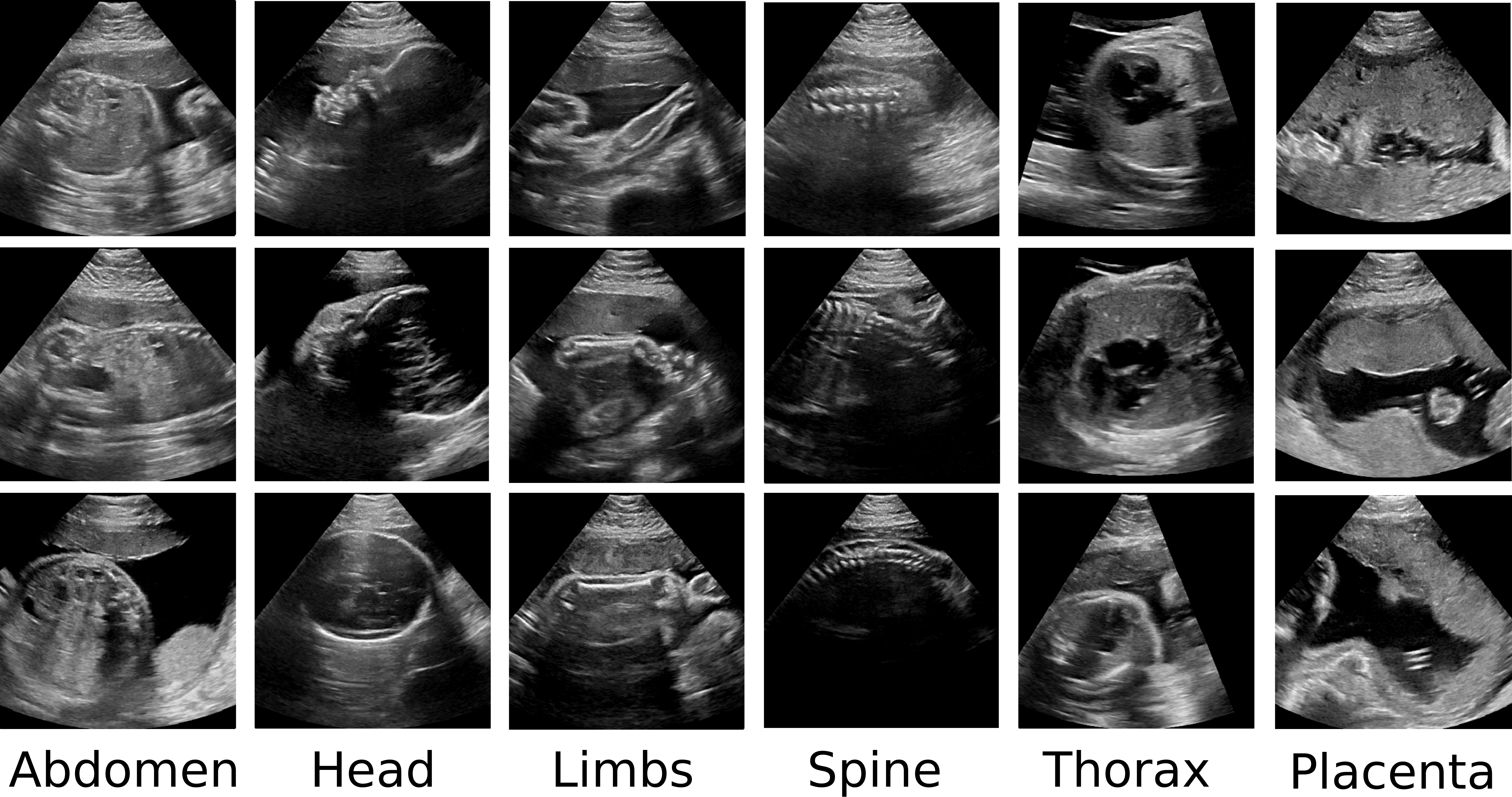}
\includegraphics[width=0.40\textwidth]{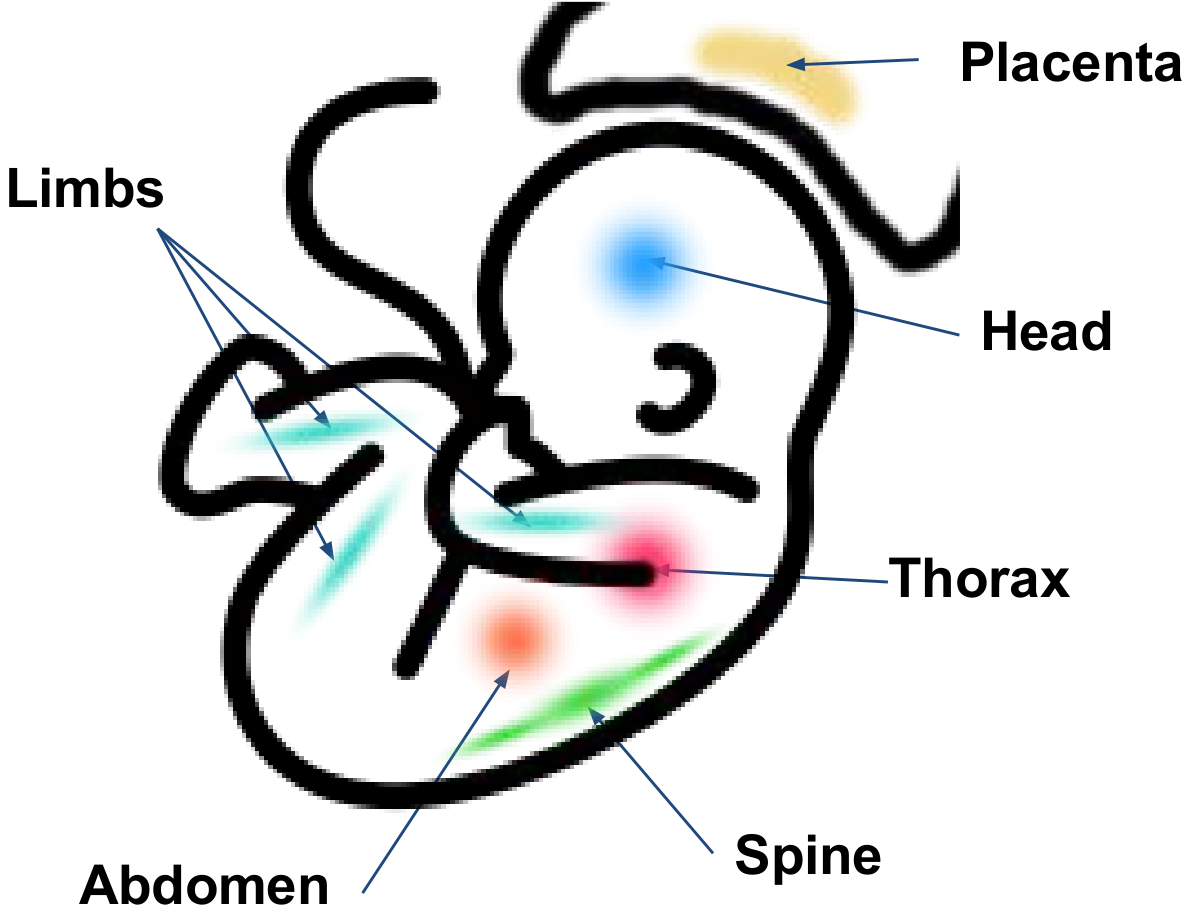}
\caption{Left: Selection of images from a fetal US examination, each column corresponds to an anatomical region. Right: Description of the anatomical regions detected and localised by our network.}
\label{fig:fetal-anatomy}
\end{figure}

\noindent \textbf{Image data: } The image data used in this work consists of a set of 20 free-hand fetal US examinations from patients, with gestational age of 27$\pm 5$ weeks from free-hand ultrasound. The system used was a Philips EPIQ 7G machine. Each examination generated a stream of approximately 40,000 frames. Example of images are shown in Fig.~\ref{fig:fetal-anatomy} (left). Each frame was stored on disk at acquisition time at full resolution and full frame rate. Acquisition parameters were provided from the manufacturer in real time.

\noindent \textbf{Image labels: } Each examination dataset was uploaded into a custom-made browser, which enabled the entire batch to be split into six different categories (\texttt{+}~background), or labels, forming an anatomical parcellation of the gestational sac. Regions are shown in Fig.~\ref{fig:fetal-anatomy} (right), and defined as follows (number of labelled frames in brackets):

\begin{itemize}
\item \textbf{Head:} [25,249 fr.] Should contain the skull, full or in part.
\item \textbf{Thorax:} [32,254 fr.] Should contain the cardiac chambers, full or in part.
\item \textbf{Abdomen:} [16,220 fr.] Should contain the abdomen (diaphragm to pelvis).
\item \textbf{Spine:} [5,980 fr.] Should contain part of the spine.
\item \textbf{Limbs:} [11,617 fr.] Should contain one or more extremity(ies).
\item \textbf{Placenta:} [6,081 fr.] Should contain part of the placenta.
\item \textbf{Background:} [12,687 fr.] No distinguishable structure in image.
\end{itemize}

\noindent Categories were chosen such that they cover the entirety of the fetal body, ensuring that any image containing fetal tissue will fall into one of them. The following heuristics were followed for categorisation:

\begin{itemize}
\item An image is categorised as label \textbf{X} if \textbf{X} is the only category visible.
\item If more than one category is visible, an image can be categorised as label \textbf{X} if \textbf{X} occupies the majority of the image.
\item Images disagreeing with those rules (indistinguishable objects(s), no prominent category, strong blur, etc) are discarded.
\end{itemize}

\noindent Labelling of the 20 datasets was performed by 3 clinical experts. The total time spent per dataset was approximately 1.5h. The labelled data was then split between training and test sets using an $80\% - 20\%$ ratio, resulting in 85,500 images for training and 24,500 for testing. The split was performed at the subject level to ensure that we were testing the generality of the network.

\section{Methods}

\subsection*{Preprocessing}

\noindent \textit{1. Polar projection:} Unlike most other imaging modalities, fetal ultrasound images are sampled in polar cordinates, yielding the characteristic frustum-shaped images. The geometric properties of the frustum vary drastically at acquisition time, depending on the organ of interest, the fetal lie or the gestational age. To prevent our network from inadvertently learning the shape of the frustum as a feature associated with a specific class (i.e. to be invariant to sector width and depth), we transformed each image into its associated polar coordinate representation. We used acquisition parameters from the frame's header in order to retrieve the intrinsic polar coordinate system (Depth of Scan, Voxel Size, Sector Width, Zoom Level).

\noindent \textit{2. Crop and resize:} In order to prevent our algorithm from focusing on acoustic reverberations artefacts, we cut $10\%$ off the polar projected image on either side in the depth direction. The resulting image was resized to a standard $224 \times 224$ pixel size.

\subsection*{Network}

We examined different architectures for the base feature extraction layers: VGG~\cite{simonyan2014very} and ResNets~\cite{he2016deep}. We used batch normalisation to accelerate convergence~\cite{ioffe2015batch}. We adapted the tail of the networks to incorporate the Soft Proposal block. As suggested in~\cite{zhu2017soft}, the soft proposal layer (SP) is inserted after the latest convolutional layer of the network. It is followed by a spatial pooling layer, and a fully connected linear classifier.

At training, our images were associated with a unique label corresponding to the prominent anatomical region present in the image. However, it is often the case in practice that multiple anatomical regions are visible in a single image. It is therefore important to consider that in the loss function used for our optimisation. We used a multi-label one \emph{vs.} all soft margin loss based on max-entropy:

\begin{equation}
L(x,y) = - \sum_i{y_i \log \frac{1}{1 + \exp(-x_i)} + (1-y_i) \log \frac{\exp(-x_i)}{1 + \exp(-x_i)}}
\end{equation}

with $x$ and $y$ the predicted and target class score vectors respectively, and $i$ the class index.

\subsection*{Region Localisation}

The objectness proposal maps from the SP-layer highlight regions of the image that were informative to the loss result $L(x,y)$, and can be used for localisation purposes. To quantify localisation accuracy, we computed a bounding box on the soft proposal map corresponding to the highest score. First the map is thresholded to 30\% above the median pixel value, and the enclosing bounding box is extracted. We compared the predicted bounding box against the one annotated by a clinical expert, using the intersection over union (IoU) metric.
\ \\

\textit{Implementation:} The labelling tool was built using C++ and Qt software. Preprocessing was performed using the Insight ToolKit. We used pyTorch for the implementation of the network architecture. We trained our networks using CUDA 8.0 on an Nvidia GeForce GTX 960M GPU.

\section{Experiments and Results}

\noindent \textbf{Training:} We trained four different feature extraction networks with batch normalisation coupled with a soft proposal layer: VGG13-SP, VGG16-SP, ResNet18-SP, and ResNet34-SP, in an end-to-end manner. We used $K=512$ feature channels in the SP layer (see~\cite{zhu2017soft}). Mini-batch Nesterov gradient descent was chosen with a momentum of 0.8. L2 regularisation was used with a weight decay of $5 \times 10^{-4}$. The initial learning rate was 0.05 and was divided by a factor 10 every 5 epochs until convergence. Since the training data contained large variability in size and orientation, the only data augmentation used was random horizontal flip. Class imbalance was addressed by weighting the probability to draw a sample by its relative class occurrence in the training set. Convergence typically occurred within ten hours.

\begin{figure}[!htb]
\includegraphics[width=\textwidth]{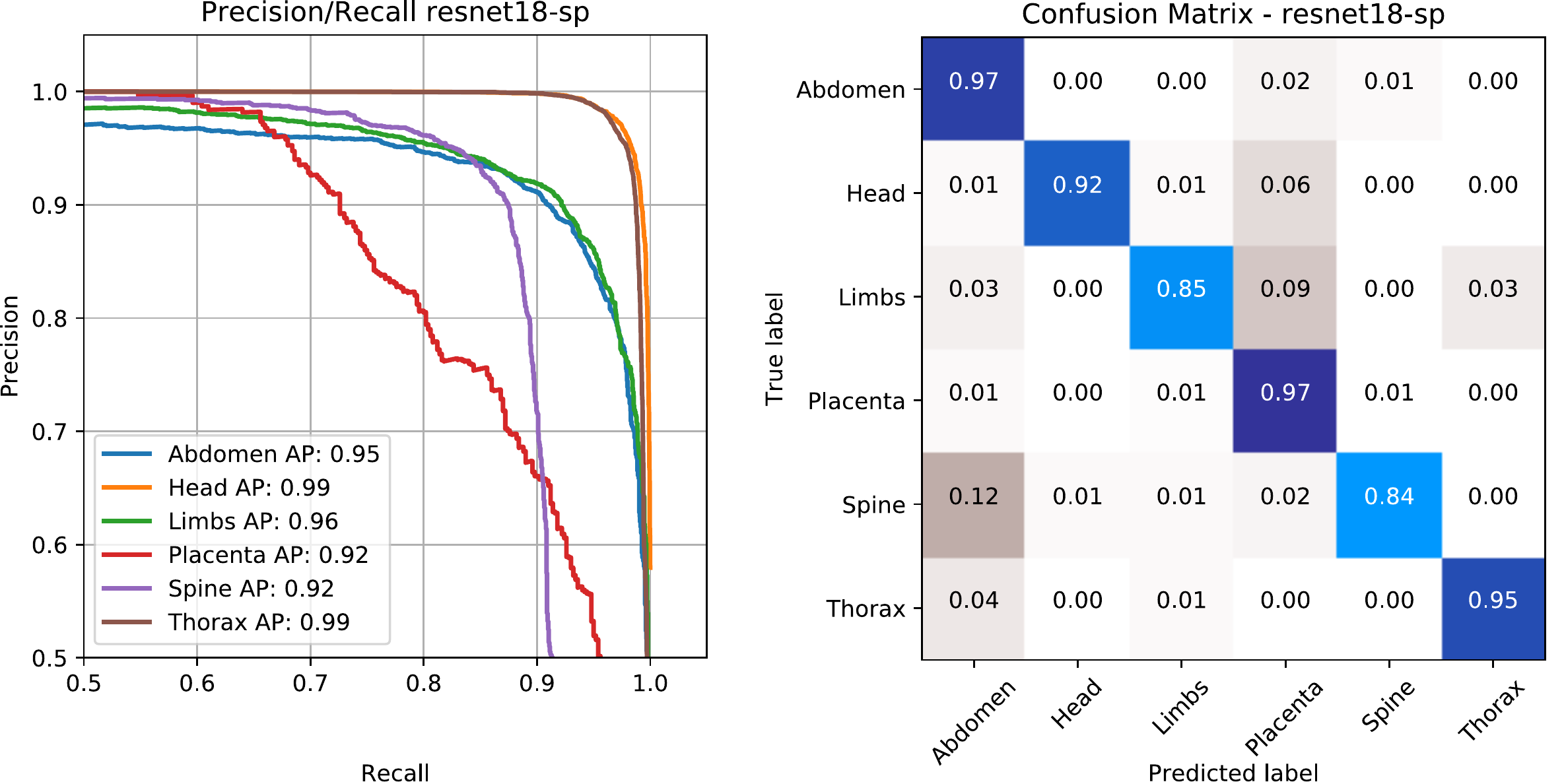}
\caption{Classification results for the ResNet18-SP network. Left: Precision/recall curves for each class. Right: Normalised confusion matrix. Figure best viewed in colour.}
\label{fig:classificationresults2}
\end{figure}

\noindent \textbf{Region detection:} After training, we evaluated the generalisation of our networks on a test set consisting of three previously unseen subjects' examinations, with a total number of 24,500 frames. The detailed classification scores of each network on the test set are summarised in Table \ref{tab:classificationresults1}. The best performing network was ResNet18-SP. To further illustrate the results of this network, we show precision/recall curves for each anatomical region and the region confusion matrix in Fig.~\ref{fig:classificationresults2}. The confusion matrix is a valuable indication on how the network is behaving in a real case scenario.

\begin{figure}[!ht]
\includegraphics[width=\textwidth]{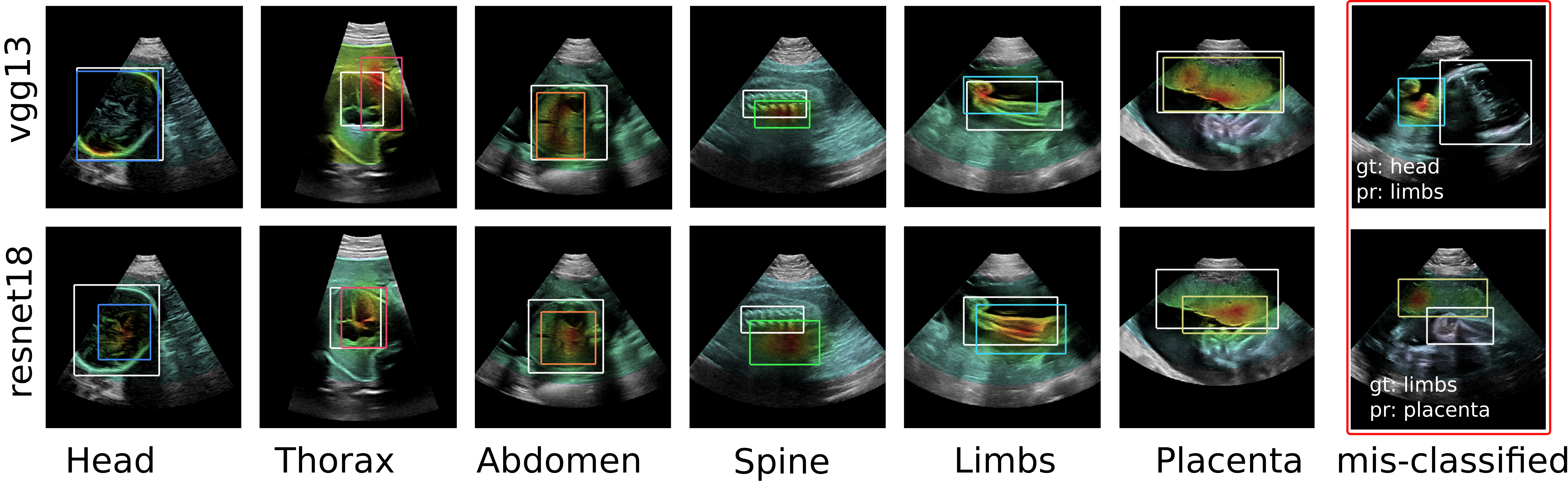}
\caption{Localisation results of VGG13-SP (top) and ResNet18-SP(bottom) for different anatomical regions. The far right column shows mis-classified examples. Images are superimposed by the soft proposal map corresponding to the predicted label. Expert bounding box is shown in white, and predicted in color. Figure best viewed in colour.}
\label{fig:localisationresults}
\end{figure}

\noindent \textbf{Region localisation:} We evaluated the correctness of the localisation task. We computed the IoU metric between predicted and ground truth bounding boxes on a randomised sub-selection of the test set, totalling 4,300 frames. The average IoU between all classes for the four architectures is reported in the last column of Table~\ref{tab:classificationresults1}. Detection and localisation resuts using ResNet18-SP and VGG13-SP for each class are shown in Fig.~\ref{fig:localisationresults}. Additionally, we illustrate in the last column some examples of mis-classifications of the networks.

\begin{table}[!htb]
\rowcolors{2}{white}{gray!15}
\centering
\begin{tabularx}{\textwidth}{l | Y | Y | Y | Y | Y | Y ? Y ? ? Y }
arch & abdo & head & limbs & plac & spine & thorax & avg & IoU \\
\hline
resnet18-sp & 0.941 & 0.922 & 0.852 & 0.968 & 0.840 & 0.947 & \textbf{0.912} & 0.393 \\
resnet34-sp & 0.945 & 0.933 & 0.686 & 0.966 & 0.871 & 0.946 & 0.891 & 0.378 \\
vgg13-sp & 0.930 & 0.921 & 0.767 & 0.978 & 0.833 & 0.899 & 0.891 & \textbf{0.424} \\
vgg16-sp & 0.948 & 0.908 & 0.785 & 0.996 & 0.839 & 0.899 & 0.896 & 0.415 \\
\end{tabularx}
\vspace{1mm}

\caption{Detailed detection scores (Accuracy) for the four SP- modified architectures on the test set for each class and their average. The last column shows the localisation scores (Intersection over Union), averaged over all classes.}
\label{tab:classificationresults1}
\end{table}

\section{Discussions}

\noindent \textbf{Detection:} Table~\ref{tab:classificationresults1} shows relatively high detection accuracy over the different regions considered. ResNet18-SP demonstrates marginally higher performance. Interestingly, deeper networks do not seem to increase performance, and can even demonstrate overfitting in the ResNet case. The limbs and spinal regions appear to be less trivial to categorise. This is partly explained by the fact that they often appear in conjunction with other regions in the field of view. The confusion matrix in Fig.~\ref{fig:classificationresults2} illustrates further this difficulty. The network can be confused between spine and abdomen, which are often seen together. A similar pattern happens between the limbs and the placenta.

\noindent \textbf{Localisation:} The last column of Table~\ref{tab:classificationresults1} reports the IoU between expert and predicted bounding boxes. While these score can appear relatively low, it is important to note that the soft proposal maps highlight regions that were discriminant for the classification task, and IoU of these order of magnitude were expected. Interestingly, the VGG backbones networks perform marginally better at agreeing with expert localisation than the ResNet ones. Fig.~\ref{fig:localisationresults} shows example of localisation results from VGG13-SP (top) and ResNet18-SP (bottom) for the different fetal regions. Both network are able to attach to anatomically relevant parts of the image. Discrepancies between soft proposal from the two networks demonstrate that, at similar classification performances, the network's attention is dependent from the internal feature extraction layer.

\noindent \textbf{Mis-classifications:} The far right column in Fig.~\ref{fig:localisationresults} shows images where the network disagreed with ground truth. They are for most cases due to the presence of mupltiple regions within the field of view. Those behaviours were expected. As a bi-product, these images further demonstrate that the network's attention is focusing on relevant anatomical regions.

\noindent \textbf{Known limitations:} Our work does not yet account for images with multiple anatomical regions at training. This situation does however occur frequently. This limitation is partly addressed in our choice of loss function but may be misleading the network in difficult cases. To fully address this issue, we will investigate the introduction of multi-labelled images at training. Another limitation of this work is the relatively small number of subjects used for training. This was however balanced by the large variability of appearances per class even within a subject, as we allow categorisation of an anatomical region from any possible angle, resulting in 9,000 images per class per subject on average.

\section{Conclusion}

In this paper, we augmented classification network architectures with soft proposal layers, and adapted them for the specific task of fetal region detection and localisation in real-time 2D ultrasound imaging. We showed that the proposed network achieves high accuracy for automatic annotation of arbitrary 2D fetal ultrasound images. Furthermore, the network is capable of localising relevant anatomical structures characteristic of each anatomical region, while there was no localisation provided at training. The ability to semantically categorise arbitrary US images could play a key role to developing navigation systems, or guide non-expert sonography scanning. Furthemore, this work could aid subsequent tasks such as scan plane detection, semantic segmentation or biometry estimation in a multi-task framework.

\noindent \textbf{Acknowledgments:} This work was supported by the Wellcome/EPSRC Centre for Medical Eng. [WT203148/Z/16/Z], and the Wellcome IEH Award [102431].

\vspace{-3mm}

\bibliographystyle{plain}

\end{document}